\documentclass{article}

\usepackage{arxiv}

\usepackage{natbib}         % author-year citations
\setcitestyle{authoryear,open={[},close={]},citesep={,},aysep={,}}
\usepackage{svg}
\usepackage[utf8]{inputenc} % allow utf-8 input
\usepackage[T1]{fontenc}    % use 8-bit T1 fonts
\usepackage[
  colorlinks=true,
  linkcolor=black,
  citecolor=black,
  urlcolor=blue
]{hyperref}       % hyperlinks
\usepackage{url}            % simple URL typesetting
\usepackage{booktabs}       % great quality tables
\usepackage{amsmath}        % math typesetting
\usepackage{amsfonts}       % blackboard math symbols
\usepackage{nicefrac}       % compact symbols for 1/2, etc.
\usepackage{microtype}      % microtypography
\usepackage{graphicx}
\usepackage{multirow}
\usepackage{appendix}
\graphicspath{{./images/}}

\title{Beam Search, Self-Consistency, and the Limits of Inference-Time Scaling for Grammar-Constrained Text-to-SQL in Small Language Models}

\author{
  Ty Chermsirivatana and John MacCormick\\
  Deep Network Understanding Lab\\
  Dickinson College\\
  \texttt{chermsit@dickinson.edu}
}

\begin{document}
\maketitle

\begin{abstract}
One common trade-off in the use of large language models involves reducing the
size of the model while increasing the amount of computation at inference time,
for example by using a wider beam search. In this paper, we examine the
\emph{constrained} case of this ``model size vs.\ inference compute'' trade-off,
in which the model outputs are constrained by a strict grammar at inference
time. Our results demonstrate that the constrained trade-off behaves differently
from the unconstrained trade-off. We investigate the task of converting a prose
query into an equivalent SQL query (text-to-SQL). Performance is evaluated on
the Spider text-to-SQL benchmark, using the Qwen2.5-Instruct model family
ranging in size from 0.5B to 7B parameters, all at 4-bit precision. We
experiment with two approaches to varying inference compute: (i)~beam search
with a variable number of beams; and (ii)~sample+vote, i.e., sampling several
constrained outputs and then voting on their execution results, where the number
of samples is varied. On the 1034-example development set, we find that:
(a)~both beam search and sample+vote improve accuracy, especially on smaller
model sizes; (b)~the ``model size vs.\ inference compute'' trade-off is not
advantageous in this experiment, because moving to a larger model size typically
results in higher accuracy than increasing inference compute on the same model
size; (c)~beam search outperforms sample+vote at a matched inference budget.
This latter result is of particular interest since it contrasts with the
findings of the \emph{unconstrained} trade-off.
\end{abstract}

\section{Introduction}
Many deployed systems require a language model to emit structured output such as
SQL, schema-conformant JSON, or function-call arguments, where the output is
useful only if it parses and runs. A common safeguard is
\emph{grammar-constrained decoding}. Each time a token is generated, any token
that would make the output inconsistent with a formal grammar is masked
\citep{geng2023gcd, willard2023outlines}. This guarantees syntactic validity
and, with a schema-aware grammar, prevents references to nonexistent tables,
columns, or objects.

Separately, \emph{inference-time scaling} spends more compute at generation time
rather than enlarging the model, for example by widening beam search or by
sampling many candidate outputs and aggregating them. The most influential
aggregation method for reasoning tasks is \emph{self-consistency} (or
\emph{sample+vote}), which samples several solutions and returns the most
frequent final answer. In unconstrained settings, this approach can
substantially outperform greedy decoding \citep{wang2022selfconsistency}.

We investigate the cost-benefit trade-off of combining these two ideas. For
example, a developer might use grammar constraints to ensure that a small model
produces valid output, then employ additional inference-time computation in an
attempt to approach the accuracy of a larger model. This strategy trades the
higher memory and per-token computational requirements of a larger model for
repeated or more extensive decoding with a smaller one. This strategy may be
attractive when the larger model exceeds the memory capacity of the available
hardware or requires a more expensive accelerator, particularly for local
deployment.

Our experiments show that this strategy has limited utility in the
grammar-constrained setting studied here. We evaluate four models from the
Qwen2.5 family (0.5B, 1.5B, 3B, and 7B parameters) at fixed 4-bit precision on
the full Spider development set of 1{,}034 examples, using execution accuracy as
the evaluation metric and comparing beam width and sampling budget at matched
cost. Our findings for this task and model family are as follows: (i)~wider
beams yield only modest improvements, and the gains shrink with model size, from
approximately $0.15$ in absolute execution accuracy for the 1.5B model to
approximately $0.05$ for the 7B model; (ii)~sample+vote never outperforms beam
search at a matched budget, contrary to prior results on unconstrained reasoning
tasks, suggesting that the benefits of self-consistency do not transfer to
grammar-constrained decoding in this setting; and (iii)~additional
inference-time computation does not in general substitute for parameters: at a
budget of eight beams,it closes between 50\% and 76\% of the gap to the next
larger model size, and only one of the three steps on the ladder, 1.5B to 3B, is
fully bridged.

\section{Related Work}
\emph{Grammar-constrained decoding} masks disallowed tokens using a context-free
grammar and an incremental parser~\citep{geng2023gcd}. Efficient implementations
are built on finite-state or stack-based machines, including regular-expression
constraints~\citep{willard2023outlines} and high-throughput grammar
engines~\citep{dong2024xgrammar}. For text-to-SQL, \emph{incremental parsing}
rejects invalid partial queries during beam search~\citep{scholak2021picard}.
\citet{park2024gad} show that the masking distorts the next-token distribution
and propose a correction. \emph{Self-consistency} samples several outputs and
selects the most consistent answer. This approach can substantially improve
output quality in unconstrained settings~\citep{wang2022selfconsistency}. In a
constrained setting such as text-to-SQL, the concept of consistency can be
generalized by regarding sampled queries as consistent if they produce the same
output when executed, even if they are structured very
differently~\citep{borchmann2025query}. We refer to this as \emph{sampling with
execution-based vote}. Beam search has a known length bias addressed by length
normalization and length penalties~\citep{wu2016gnmt}, tuned per-token
rewards~\citep{murray2018length}, and bounded rewards~\citep{yang2018breaking}.
A related line of research studies the end-of-sequence decision and
termination~\citep{newman2020eos, welleck2020consistency, kasai2022beam}. In our
setting, the schema-aware grammar never forces a query to end, and the
termination decision is thus in the hands of the model and/or the beam search
algorithm; we focus on measuring the accuracy rather than controlling the length
of the response.

\section{Problem Formulation}
Consider an autoregressive language model over a vocabulary $V$, which generates
a token sequence $y = (y_1, \dots, y_t)$. Let $G$ be a context-free grammar,
which defines the set of valid strings $L(G)$. \emph{Grammar-constrained
decoding} restricts a newly generated token $y_t$ to the set $A_G(y_{<t})
\subseteq V$ of tokens that keep the prefix $y_{\le t}$ extendable to some
string in $L(G)$. The model thus decodes from the masked distribution that
zeroes every token outside $A_G(y_{<t})$ and renormalizes. The end-of-sequence
token is admitted only at an accepting state, that is, when $y_{<t} \in L(G)$. A
schema-aware grammar further restricts identifiers to the tables and columns
present in the target database $D$.

Two ways of spending a computational budget $B$ on this grammar-constrained
model are compared. \emph{Beam search} maintains $B$ partial hypotheses, expands
each at every step, and retains the $B$ highest-scoring continuations. A
hypothesis that emits the end-of-sequence token leaves the beam and joins a set
of finished candidates scored with a length penalty, and decoding halts once
that set holds $B$ candidates, returning the highest-scoring one. This
first-come, first-served stopping rule is the one implemented by the library used
here and is analyzed by \citet{kasai2022beam}. The length penalty makes finished
queries preferable to longer continuations~\citep{wu2016gnmt, murray2018length}.
This penalty is particularly desirable when $L(G)$ always permits a query to
continue, as in our experiment. \emph{Sampling with execution-based vote}
(abbreviated to \emph{sample+vote}) draws $B$ independent samples at
temperature $\tau$, executes each against $D$, discards those that fail, and
returns a query that produced the modal result $\hat{r} =
\operatorname{mode}\{r^{(1)}, \dots, r^{(B)}\}$. The two are compared at matched
budget $B \in \{1, 2, 4, 8\}$, where $B = 1$ is greedy decoding for beam search
and a single draw for sampling.

A natural question is whether sample+vote is merely a noisy variant of beam
search. In the limit $\tau \to 0$, sample+vote collapses to greedy decoding,
which is beam search with $B = 1$, so the two methods coincide and the
comparison is uninformative. The comparison is therefore meaningful only at
temperatures high enough to produce diverse samples. We use $\tau = 0.7$, the
standard choice in the self-consistency
literature~\citep{wang2022selfconsistency}. A full temperature sweep, which we
expect to interpolate between the two methods, is left to future work.

A prediction $\hat{y}$ is \emph{correct} when it and the gold query $y^{*}$
return the same result set on $D$, with order significant only when $y^{*}$
contains an \texttt{ORDER BY} clause. \emph{Execution accuracy} is the fraction
of correct predictions. Write $\operatorname{acc}(s, B)$ for the execution
accuracy of a size-$s$ model at budget $B$. Then the compute-vs-parameters
question asks whether $\operatorname{acc}(s, B) \ge \operatorname{acc}(s', 1)$
for some size $s < s'$ and some feasible $B$.

\section{Experimental Setup}
\label{sec:setup}
Our evaluation employs the Qwen2.5-Instruct family of models~\citep{qwen2024},
with sizes $s$ at 0.5B, 1.5B, 3B, and 7B parameters. All experiments employ
4-bit (NF4) quantization. Every model and method is evaluated on the full
development set of 1{,}034 examples from the Spider
benchmark~\citep{yu2018spider}. Hence, each execution accuracy measurement is a
binomial proportion over a population of the same size ($n = 1034$), and the
associated confidence intervals are directly comparable across configurations.
Sample+vote uses temperature $\tau = 0.7$ with nucleus sampling at top-$p =
0.9$, and beam search uses a length penalty of $-2.0$, chosen so that finished
queries are preferred over longer continuations. Generation is capped at 160 new
tokens.

Each configuration is evaluated in a single decoding run over the 1{,}034
examples. Beam search is deterministic, and thus has no dependence on the random
number seed. In contrast, outputs of the sample+vote approach do depend on the
random number seed. However, our experiment comprised a single run with a single
seed; an analysis of intrinsic variability in sample+vote is left to future
work. We do consider another source of uncertainty, which is the selection of a
particular population of test examples by the Spider benchmark. The results
below show error bars representing 95\% Wilson score intervals for a binomial
proportion with $n = 1034$, quantifying the uncertainty over test examples
rather than variability across repeated sampling runs.

Two kinds of uncertainty are therefore reported, and they answer different
questions. The error bars just described concern how precisely each individual
accuracy is estimated. They are not a test of the difference between two
accuracies: two intervals can overlap while the underlying difference is
significant, so comparisons in the text are not made by inspecting overlap.
Because every configuration is evaluated on the same 1{,}034 examples, any two
configurations are \emph{paired}, and the appropriate test is McNemar's exact
test, which conditions on the examples where the two configurations disagree and
ignores the much larger number on which they agree. All statements regarding
statistical significance in this paper use that test.

\section{Results}

\begin{table}[t]
  \caption{\textbf{Execution accuracy on the Spider development set ($n = 1034$) by model size, method, and budget.} The no-grammar greedy baseline is shown for reference. Each cell is a single decoding run over the full set.}
  \centering
  \begin{tabular}{llccccc}
    \toprule
    Model & No-grammar greedy & Method & Budget 1 & Budget 2 & Budget 4 & Budget 8 \\
    \midrule
    \multirow{2}{*}{Qwen2.5-0.5B} & \multirow{2}{*}{0.135} & Beam          & 0.139 & 0.203 & 0.238 & 0.245 \\
                                  &                        & Sample+vote   & 0.113 & 0.159 & 0.201 & 0.248 \\
    \midrule
    \multirow{2}{*}{Qwen2.5-1.5B} & \multirow{2}{*}{0.326} & Beam          & 0.351 & 0.482 & 0.502 & 0.505 \\
                                  &                        & Sample+vote   & 0.299 & 0.348 & 0.403 & 0.450 \\
    \midrule
    \multirow{2}{*}{Qwen2.5-3B}   & \multirow{2}{*}{0.477} & Beam          & 0.445 & 0.516 & 0.549 & 0.562 \\
                                  &                        & Sample+vote   & 0.441 & 0.469 & 0.489 & 0.513 \\
    \midrule
    \multirow{2}{*}{Qwen2.5-7B}   & \multirow{2}{*}{0.643} & Beam          & 0.600 & 0.648 & 0.662 & 0.654 \\
                                  &                        & Sample+vote   & 0.601 & 0.615 & 0.628 & 0.640 \\
    \bottomrule
  \end{tabular}
  \label{tab:main}
\end{table}

\begin{figure}[t]
  \centering
  \includegraphics[width=0.85\textwidth]{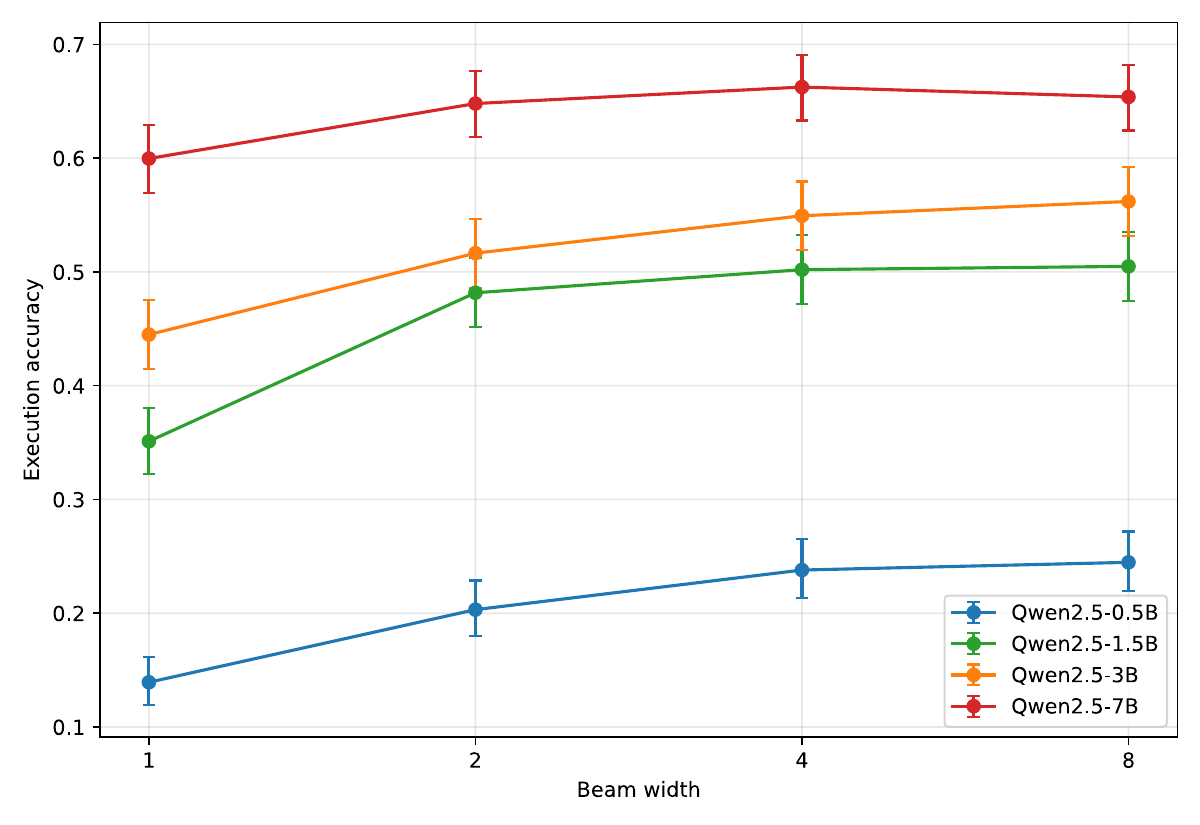}
  \caption{\textbf{Execution accuracy versus beam width for each model size.}
  The benefit of wider beams decreases with size and saturates by width~4. Error
  bars are 95\% Wilson score intervals over the $n = 1034$ test examples, and
  each point is one decoding run (beam search is deterministic).}
  \label{fig:beamwidth}
\end{figure}

\begin{figure}[t]
  \centering
  \includegraphics[width=0.95\textwidth]{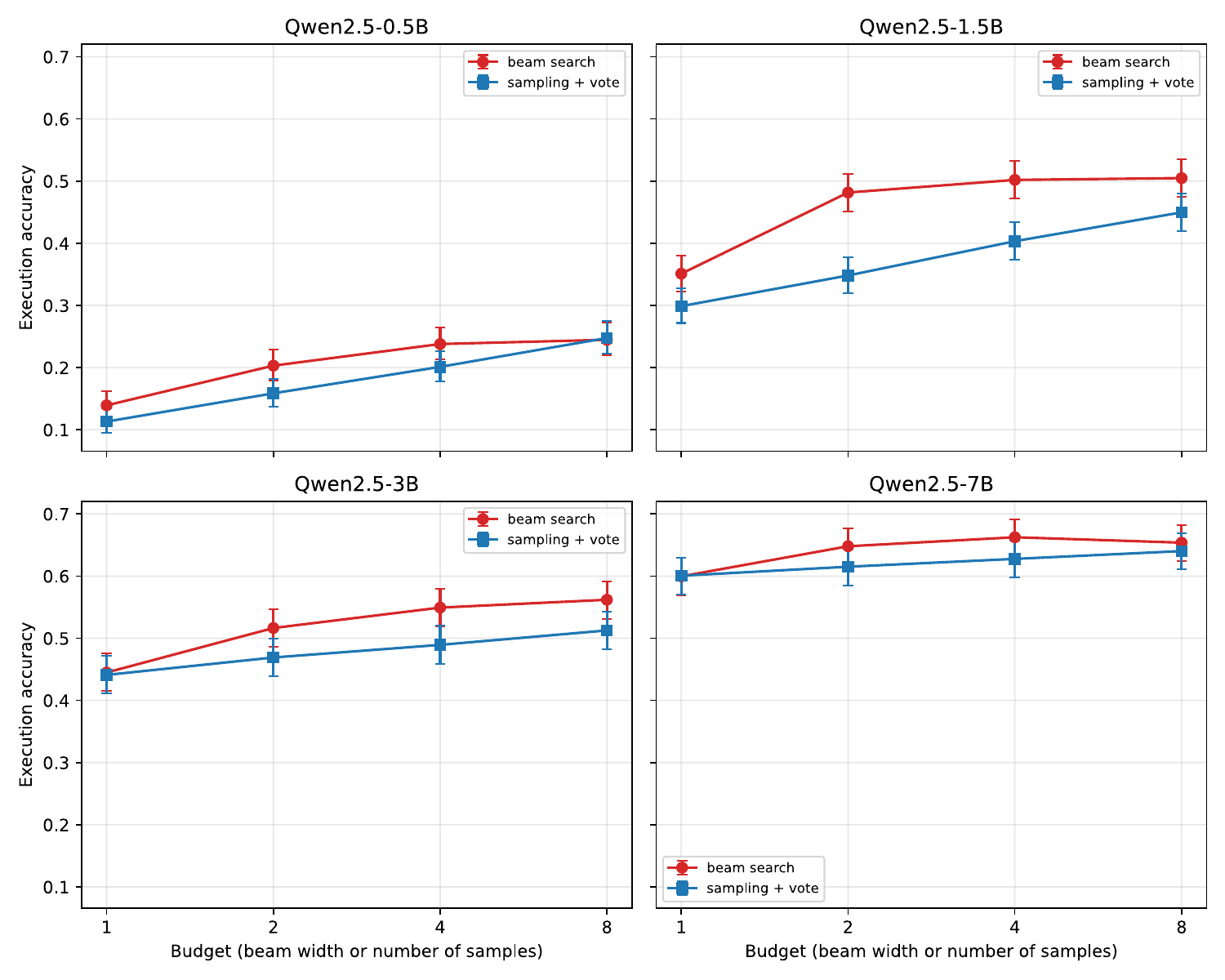}
  \caption{\textbf{Beam search compared with sample+vote at matched budget.}
  Sample+vote is never significantly more accurate than beam search. Beam
  search is significantly ahead in 11 of the 16 cells by an exact McNemar test
  on the paired predictions (see Table~\ref{tab:mcnemar}) and significantly
  behind in none; the remaining five cells are statistical ties. Error bars are
  95\% Wilson score intervals for the individual accuracies. As discussed in
  Section~\ref{sec:setup}, each point represents a single decoding run
  (employing a single random number seed for sample+vote).}
  \label{fig:beamvssampling}
\end{figure}

\paragraph{The grammar constraint itself can cost accuracy.}
Table~\ref{tab:main} shows that at 3B and 7B the \emph{unconstrained} greedy
baseline beats grammar-constrained decoding at budget 1 (0.477 vs.\ 0.445 and
0.643 vs.\ 0.600). Since budget-1 beam search is greedy decoding under the mask,
the gap is attributable to the constraint rather than to the search. It is not,
however, attributable to distributional distortion. Masking followed by
renormalization rescales all admitted tokens by a common factor and therefore
cannot reorder them, so whenever the unconstrained greedy output already lies in
$L(G)$, constrained greedy decoding must reproduce it exactly. The data are
consistent with this. The two decoders emit identical strings on 60.4\% of
examples at 3B and 52.5\% at 7B, and the deficit is confined entirely to the
examples on which they differ: at 3B the constraint corrects 25 predictions and
breaks 58, and at 7B it corrects 28 and breaks 73, both significant under an
exact McNemar test on the paired predictions ($p = 3.8\times10^{-4}$ and $p =
8.6\times10^{-6}$).

Re-parsing each differing prediction under the grammar identifies the mechanism
as incomplete coverage. At 3B every one of the 400 testable differing
predictions is rejected by our grammar, and each rejected prediction that was
nonetheless correct parses as valid SQLite under an independent parser. These
correct-but-rejected predictions executed successfully against the gold
database, so these are constructs the grammar cannot derive, rather than
malformed output. In other words, the particular grammar employed for these
experiments is \emph{incomplete}. The recurring cases of incompleteness are:
value lists in \texttt{IN}, for which our grammar admits only a subquery and so
forces a degenerate repetition; \texttt{IS NOT NULL}, which it cannot express;
outer joins; and free-form or abbreviated table aliases. Each rejection deflects
generation mid-query, after which the forced continuation can be globally
unlikely in the manner analyzed by \citet{park2024gad}. This also explains the
trend with model size. A stronger model writes idiomatic SQL that is already
valid, so the grammar is less likely to help, but the incompleteness of the
grammar still imposes a penalty. A smaller model size such as 1.5B needs more
help from the grammar. In this case, the grammar produces a statistically
significant net gain (0.351 vs.\ 0.326, $p = 0.023$).

\paragraph{Inference-time compute significantly improves accuracy, especially for smaller model sizes.}
The general trend in Figures~\ref{fig:beamwidth} and~\ref{fig:beamvssampling}
and Table~\ref{tab:main} is clear: for each fixed model size examined here,
increasing the inference-time compute budget \(B\) improves execution accuracy.
These benefits are concentrated among the smaller model sizes. The
\(0.5\mathrm{B}\)- and \(1.5\mathrm{B}\)-parameter models achieve a
\(1.2\times\) to \(1.4\times\) boost in accuracy when \(B\) increases from \(1\)
to \(2\), rising to a \(1.5\times\) to \(2.2\times\) boost when \(B=8\). In
contrast, the \(3\mathrm{B}\)- and \(7\mathrm{B}\)-parameter models receive more
modest accuracy boosts, ranging from \(1.1\times\) to \(1.3\times\) when
\(B=8\). For beam search, the accuracy boost appears to saturate around \(B=4\)
to~\(8\). In contrast, the sample+vote method does not appear to saturate at the
maximum value of \(B=8\) tested here, especially for the two smaller model sizes
(see Figure~\ref{fig:beamvssampling}).

\begin{table}[htbp]
  \centering
  \caption{\textbf{Paired comparison of beam search and sample+vote.} The
  table shows the significance of the difference between each pair of points in
  Figure~\ref{fig:beamvssampling}. Because beam search and sample+vote are
  both evaluated on the same $n=1034$ examples, the appropriate test is
  McNemar's exact test on the discordant pairs rather than an interval for two
  independent proportions. ``Beam only'' counts examples beam answered correctly
  and sampling did not, and conversely for ``Sampling only''; ``Agree'' counts
  examples on which the two methods gave the same verdict. Asterisks mark $p <
  0.05$.}
  \label{tab:mcnemar}
  \begin{tabular}{llcccccc}
    \toprule
          &     & \multicolumn{2}{c}{Accuracy} & \multicolumn{2}{c}{Correct examples} &       &     \\
    Model & $B$ & Beam        & Sampling       & Beam only    & Sampling only         & Agree & $p$ \\
    \midrule
    \multirow{4}{*}{Qwen2.5-0.5B}
      & 1 & 0.139 & 0.113 & 73  & 46 & 915 & $0.017^{*}$ \\
      & 2 & 0.203 & 0.159 & 103 & 57 & 874 & $3.4\times10^{-4\,*}$\\
      & 4 & 0.238 & 0.201 & 115 & 77 & 842 & $0.0074^{*}$ \\
      & 8 & 0.245 & 0.248 & 95  & 98 & 841 & $0.89$ \\
    \midrule
    \multirow{4}{*}{Qwen2.5-1.5B}
      & 1 & 0.351 & 0.299 & 116 & 62 & 856 & $6.3\times10^{-5\,*}$\\
      & 2 & 0.482 & 0.348 & 204 & 66 & 764 & $1.4\times10^{-17\,*}$\\
      & 4 & 0.502 & 0.403 & 183 & 81 & 770 & $3.1\times10^{-10\,*}$\\
      & 8 & 0.505 & 0.450 & 154 & 97 & 783 & $3.9\times10^{-4\,*}$\\
    \midrule
    \multirow{4}{*}{Qwen2.5-3B}
      & 1 & 0.445 & 0.441 & 57  & 53 & 924 & $0.78$ \\
      & 2 & 0.516 & 0.469 & 104 & 55 & 875 & $1.3\times10^{-4\,*}$\\
      & 4 & 0.549 & 0.489 & 113 & 51 & 870 & $1.5\times10^{-6\,*}$\\
      & 8 & 0.562 & 0.513 & 122 & 71 & 841 & $3.0\times10^{-4\,*}$\\
    \midrule
    \multirow{4}{*}{Qwen2.5-7B}
      & 1 & 0.600 & 0.601 & 32 & 33 & 969 & $1.00$ \\
      & 2 & 0.648 & 0.615 & 67 & 33 & 934 & $8.7\times10^{-4\,*}$\\
      & 4 & 0.662 & 0.628 & 73 & 37 & 924 & $7.7\times10^{-4\,*}$\\
      & 8 & 0.654 & 0.640 & 65 & 51 & 918 & $0.23$ \\
    \bottomrule
  \end{tabular}
\end{table}

\paragraph{Sampling with voting never beats beam search at a matched budget.}
The sixteen McNemar test results in Table~\ref{tab:mcnemar} show that beam
search is significantly more accurate than sample+vote in eleven and
significantly less accurate in none. The advantage is largest at 1.5B.

Sample+vote does improve over greedy decoding at every model size here,
confirming previous results in the self-consistency literature, such as Wang et
al.~(\citeyear{wang2022selfconsistency}). However, our results contrast with Wang
et al.\ when comparing sample+vote against beam search at a matched budget.
Wang et al.\ found that sampling beats beam search on reasoning tasks; they
attributed this to the higher diversity of sampled outputs. In contrast, for the
task evaluated here, beam search is superior to sample+vote. We cannot
generalize confidently from the single text-to-SQL task evaluated here, but one
might hypothesize that beam search has a general advantage over sample+vote in
constrained settings.

\paragraph{In two of the three cases tested, inference compute fails to substitute for parameters.}
Our main research question asks whether increased inference compute can
compensate for smaller model size. Figure~\ref{fig:beamwidth} shows that the
answer is negative for two of the three cases tested: an $8\times$ increase in
inference compute does \emph{not} compensate for a downgrade from 7B to 3B
parameters, and similarly for 1.5B to 0.5B. In one case however, we do see an
acceptable trade-off: a $2\times$ (i.e., $B=2$) inference compute investment
does appear to compensate for a downgrade from 3B for 1.5B.

\section{Discussion}
Why does beam search perform better than sample+vote for our text-to-SQL task
at matched budgets?
One natural hypothesis is the capability for \emph{over-generation}: the grammar always
permits a query to continue, so beam search can extend queries past a valid
stopping point. The data do not support this hypothesis. With a length penalty of $-2.0$,
the truncation rate \emph{falls} monotonically with beam width, from between
0.8\% and 2.4\% at width 1 to between 0.0\% and 1.1\% at width 8. The truncation rate reaches zero
at 3B and 7B, because a wider beam has more opportunities to find a completed
hypothesis and the penalty rewards finishing. Sample+vote truncates between 1.0\%
and 2.7\% at every size, so beam search over-generates no more than sample+vote
does at any width above 1. Over-generation therefore does not explain the
superior performance of beam search. This also indicates that a negative length penalty is sufficient to
control non-termination in grammar-constrained beam search, a setting where the
grammar itself never forces a query to end~\citep{newman2020eos,
welleck2020consistency, kasai2022beam}. Two other hypotheses seem plausible.
First, the grammar removes most of the surface diversity that self-consistency
exploits, since every candidate is a valid query over the same schema. Second, a
weak constrained model appears to produce samples that agree on a confident but
wrong result, so sample+vote concentrates on that result rather than averaging out
errors. Self-consistency needs diverse, approximately unbiased samples;
under a hard grammar constraint, neither condition seems to hold. Verifying these
mechanisms directly, for example by measuring sample diversity and per-example
vote entropy, is left to future work.

Future work should also examine the impact of size of the dataset. Our
exploratory experiments, not reported here, found that evaluations on
\emph{subsets} of the development set generally favored sample+vote over beam
search. This effect vanished only at the full $n = 1034$. Hence, it seems
text-to-SQL evaluations on small datasets could produce misleading results.

\section{Limitations}

Our experiments were limited in scope to one model family (Qwen2.5-Instruct),
one benchmark (Spider), and a single decoding run per configuration. Therefore,
the above results suggest plausible conjectures about more general properties of
constrained LLMs, but we cannot form general conclusions without a much more
diverse set of experiments.

\section{Conclusion}
We compared beam search to sample+vote in a grammar-constrained text-to-SQL
task, in which the LLM was permitted to output only valid SQL. Experiments were
conducted using the Qwen2.5-Instruct family of LLMs, from 0.5B to 7B parameters,
applied to the Spider development set. Our most striking result was that, in our
\emph{grammar-constrained} setting, beam search was superior to sample+vote.
This contrasted with previous work showing that in an \emph{unconstrained}
setting, sample+vote performs better than beam search. We also found that, in
the grammar-constrained setting, inference-time compute (whether via beam search
or sample+vote) significantly improved accuracy, especially for the smaller
model sizes. A third important result was that inference-time compute typically
did not compensate for reduced model size. The most direct avenue for future
work is to test whether these three results persist across other model families,
other constraint types such as JSON schemas and tool-call grammars, other
benchmarks, and repeated sampling runs.

\section*{Author Contributions}
Chermsirivatana conceived the idea, designed and ran the experiments, and wrote
the first draft of the paper. MacCormick advised on interpretation of the
results and edited the draft. AI assistants were used for grammar and style
checking, and for some content generation at the sentence level.

\bibliographystyle{unsrtnat}

\begin{thebibliography}{15}

\bibitem[Borchmann and Wydmuch(2025)]{borchmann2025query}
{\L}ukasz Borchmann and Marek Wydmuch.
\newblock Query and conquer: Execution-guided {SQL} generation.
\newblock \emph{arXiv preprint arXiv:2503.24364}, 2025.

\bibitem[Dong et~al.(2025)Dong, Ruan, Cai, Lai, Xu, Zhao, and Chen]{dong2024xgrammar}
Yixin Dong, Charlie~F. Ruan, Yaxing Cai, Ruihang Lai, Ziyi Xu, Yilong Zhao, and Tianqi Chen.
\newblock {XGrammar}: Flexible and efficient structured generation engine for large language models.
\newblock In \emph{Proceedings of the 7th Annual Conference on Machine Learning and Systems (MLSys)}, 2025.
\newblock Also available as arXiv:2411.15100.

\bibitem[Geng et~al.(2023)Geng, Josifoski, Peyrard, and West]{geng2023gcd}
Saibo Geng, Martin Josifoski, Maxime Peyrard, and Robert West.
\newblock Grammar-constrained decoding for structured {NLP} tasks without finetuning.
\newblock In \emph{Proceedings of the 2023 Conference on Empirical Methods in Natural Language Processing (EMNLP)}, pages 674--686, 2023.
\newblock Also available as arXiv:2305.13971.

\bibitem[Kasai et~al.(2024)Kasai, Sakaguchi, Le~Bras, Radev, Choi, and Smith]{kasai2022beam}
Jungo Kasai, Keisuke Sakaguchi, Ronan Le~Bras, Dragomir Radev, Yejin Choi, and Noah~A. Smith.
\newblock A call for clarity in beam search: How it works and when it stops.
\newblock In \emph{Proceedings of the 2024 Joint International Conference on Computational Linguistics, Language Resources and Evaluation (LREC-COLING)}, pages 77--90, 2024.
\newblock Also available as arXiv:2204.05424.

\bibitem[Murray and Chiang(2018)]{murray2018length}
Kenton Murray and David Chiang.
\newblock Correcting length bias in neural machine translation.
\newblock In \emph{Proceedings of the Third Conference on Machine Translation (WMT)}, pages 212--223, 2018.
\newblock Also available as arXiv:1808.10006.

\bibitem[Newman et~al.(2020)Newman, Hewitt, Liang, and Manning]{newman2020eos}
Benjamin Newman, John Hewitt, Percy Liang, and Christopher~D. Manning.
\newblock The {EOS} decision and length extrapolation.
\newblock In \emph{Proceedings of the Third BlackboxNLP Workshop on Analyzing and Interpreting Neural Networks for NLP (EMNLP)}, 2020.
\newblock Also available as arXiv:2010.07174.

\bibitem[Park et~al.(2024)Park, Wang, Berg-Kirkpatrick, Polikarpova, and D'Antoni]{park2024gad}
Kanghee Park, Jiayu Wang, Taylor Berg-Kirkpatrick, Nadia Polikarpova, and Loris D'Antoni.
\newblock Grammar-aligned decoding.
\newblock In \emph{Advances in Neural Information Processing Systems (NeurIPS)}, 2024.
\newblock Also available as arXiv:2405.21047.

\bibitem[Qwen Team(2024)]{qwen2024}
Qwen Team.
\newblock Qwen2.5 technical report.
\newblock \emph{arXiv preprint arXiv:2412.15115}, 2024.

\bibitem[Scholak et~al.(2021)Scholak, Schucher, and Bahdanau]{scholak2021picard}
Torsten Scholak, Nathan Schucher, and Dzmitry Bahdanau.
\newblock {PICARD}: Parsing incrementally for constrained auto-regressive decoding from language models.
\newblock In \emph{Proceedings of the 2021 Conference on Empirical Methods in Natural Language Processing (EMNLP)}, pages 9895--9901, 2021.
\newblock Also available as arXiv:2109.05093.

\bibitem[Wang et~al.(2023)Wang, Wei, Schuurmans, Le, Chi, Narang, Chowdhery, and Zhou]{wang2022selfconsistency}
Xuezhi Wang, Jason Wei, Dale Schuurmans, Quoc~V. Le, Ed~H. Chi, Sharan Narang, Aakanksha Chowdhery, and Denny Zhou.
\newblock Self-consistency improves chain of thought reasoning in language models.
\newblock In \emph{Proceedings of the Eleventh International Conference on Learning Representations (ICLR)}, 2023.
\newblock Also available as arXiv:2203.11171.

\bibitem[Welleck et~al.(2020)Welleck, Kulikov, Kim, Pang, and Cho]{welleck2020consistency}
Sean Welleck, Ilia Kulikov, Jaedeok Kim, Richard~Yuanzhe Pang, and Kyunghyun Cho.
\newblock Consistency of a recurrent language model with respect to incomplete decoding.
\newblock In \emph{Proceedings of the 2020 Conference on Empirical Methods in Natural Language Processing (EMNLP)}, pages 1819--1827, 2020.
\newblock Also available as arXiv:2002.02492.

\bibitem[Willard and Louf(2023)]{willard2023outlines}
Brandon~T. Willard and R{\'e}mi Louf.
\newblock Efficient guided generation for large language models.
\newblock \emph{arXiv preprint arXiv:2307.09702}, 2023.

\bibitem[Wu et~al.(2016)]{wu2016gnmt}
Yonghui Wu et~al.
\newblock Google's neural machine translation system: Bridging the gap between human and machine translation.
\newblock \emph{arXiv preprint arXiv:1609.08144}, 2016.

\bibitem[Yang et~al.(2018)Yang, Huang, and Ma]{yang2018breaking}
Yilin Yang, Liang Huang, and Mingbo Ma.
\newblock Breaking the beam search curse: A study of (re-)scoring methods and stopping criteria for neural machine translation.
\newblock In \emph{Proceedings of the 2018 Conference on Empirical Methods in Natural Language Processing (EMNLP)}, pages 1832--1843, 2018.
\newblock Also available as arXiv:1808.09582.

\bibitem[Yu et~al.(2018)Yu, Zhang, Yang, Yasunaga, Wang, Li, Ma, Li, Yao, Roman, Zhang, and Radev]{yu2018spider}
Tao Yu, Rui Zhang, Kai Yang, Michihiro Yasunaga, Dongxu Wang, Zifan Li, James Ma, Irene Li, Qingning Yao, Shanelle Roman, Zilin Zhang, and Dragomir Radev.
\newblock Spider: A large-scale human-labeled dataset for complex and cross-domain semantic parsing and text-to-{SQL} task.
\newblock In \emph{Proceedings of the 2018 Conference on Empirical Methods in Natural Language Processing (EMNLP)}, pages 3911--3921, 2018.
\newblock Also available as arXiv:1809.08887.

\end{thebibliography}

\end{document}